\documentclass{article}
\usepackage{spconf,amsmath,graphicx,hyperref}

\usepackage{booktabs}   % for nice table rules
\usepackage{multirow}   % for multirow cells
\usepackage{caption}    % for caption formatting
\usepackage{adjustbox}  % optional, for fitting wide tables
\usepackage{siunitx}    % for number alignment
\usepackage[table]{xcolor}
\usepackage{graphicx}
\apptocmd{\thebibliography}{\setlength{\itemsep}{0.9pt}\setlength{\parskip}{0.8pt}\linespread{0.9}\selectfont}{}{}

\title{Visual Language Model as a Judge for Object Detection in Industrial Diagrams}
\name{Sanjukta Ghosh}
\address{Siemens AG, Erlangen, Germany}
\begin{document}
%\ninept
%
\maketitle
\begin{abstract}
Industrial diagrams such as piping and instrumentation diagrams (P\&IDs) are essential for the design, operation, and maintenance of industrial plants. Converting these diagrams into digital form is an important step toward building digital twins and enabling intelligent industrial automation. A central challenge in this digitalization process is accurate object detection. Although recent advances have significantly improved object detection algorithms, there remains a lack of methods to automatically evaluate the quality of their outputs. This paper addresses this gap by introducing a framework that employs Visual Language Models (VLMs) to assess object detection results and guide their refinement. The approach exploits the multimodal capabilities of VLMs to identify missing or inconsistent detections, thereby enabling automated quality assessment and improving overall detection performance on complex industrial diagrams.

\end{abstract}
\section{Introduction}
\label{sec:intro}

While object detection has advanced significantly, its application to industrial diagrams like P\&IDs presents unique challenges related to reliability and certainty. The analysis of these diagrams underpins critical design, engineering, operations, and maintenance tasks, necessitating an extremely high degree of accuracy. Consequently, the outputs of automated detection systems must be rigorously assessed to ensure high-quality results. A common approach, known as human-in-the-loop validation, is often employed. However, this method can impose a substantial cognitive load on human operators, potentially diminishing or even negating the efficiency gains of automated systems. To address this critical limitation, a novel approach is proposed that uses a Visual Language Model (VLM) as a judge to automatically assess object detection outputs by leveraging its multi-modal capabilities. The proposed framework uses a VLM as a high-level semantic assessor, coupled with localized VLM analysis, to identify missing or inconsistent detections from a base object detector.

\section{Related Work}
\label{sec:relatedwork}
The Transformer architecture \cite{vaswani2017attention} transformed deep learning by enabling efficient modeling of long-range dependencies through self-attention. It quickly became the foundation of Natural Language Processing (NLP), powering models such as BERT \cite{devlin2018bert} and the GPT series \cite{radford2018improving,radford2019language,brown2020language} that demonstrated unprecedented scalability and transfer learning. This success inspired the application of Transformers to vision, notably the Vision Transformer (ViT) \cite{dosovitskiy2021an}, which treats images as patch sequences and rivals Convolutional Neural Networks (CNNs) \cite{alex,simonyan,kaiming} on image classification. Subsequent variants such as DeiT \cite{touvron2021training}, Swin Transformer \cite{liu2021swin}, PVT \cite{wang2021pyramid}, SAM \cite{Kirillov} improved data efficiency and enabled dense prediction tasks, cementing Transformers as a core vision backbone.

Building on these advances, Vision-Language Models (VLMs) emerged to jointly learn visual–textual representations. Large-scale models like CLIP \cite{radford2021learning}, ALIGN \cite{jia2021scaling}, BLIP \cite{li2022blip}, SigLIP \cite{siglip} and Flamingo \cite{alayrac2022flamingo} leverage massive image–text corpora to achieve strong zero-shot and few-shot performance on tasks like visual question answering, captioning, retrieval, visual reasoning \cite{vr1}. 

PFD and P\&ID digitalization has been addressed using deep learning techniques, such as CNNs and transformers \cite{paliwal,gajbhiye,kim, theisen, tresp,sturmer_2025_14803338}.

\section{Problem Description}
\label{sec:problem}
\begin{figure}[!htbp]
  \centering
  \includegraphics[width=\columnwidth]{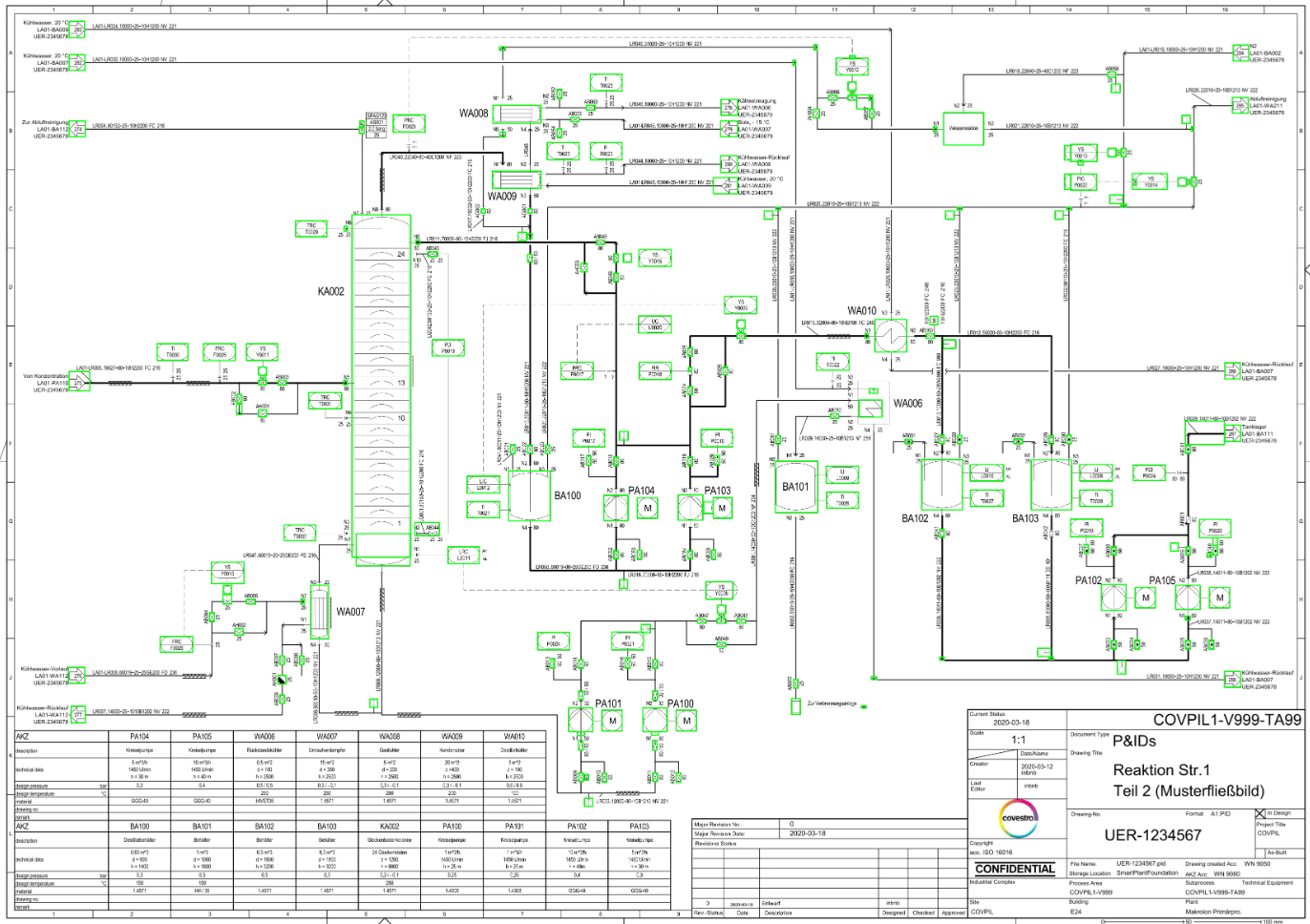}
  \caption{Typical P\&ID detection results.}
  \label{fig:pic1}
\end{figure}

Detecting symbols in industrial diagrams like P\&IDs is a critical step towards their digitalization. However, given that these diagrams are large, dense and intricate as shown in Figure~\ref{fig:pic1}, it is often observed that conventional object detectors tend to miss certain objects. Sometimes the same object detectors are iterated upon to improve detection rates; however, this is not always effective. Given the need for high accuracy in industrial applications, it is crucial to assess the quality of object detection outputs.

The validation process can be systematically broken down into four key aspects:
1) Completeness: Assessing whether any relevant objects have been missed by the detection model.
2) Precision: Evaluating the number of false detections, i.e., instances where the model incorrectly identifies an object.
3) Localization Accuracy: Determining the precision of the bounding box coordinates for each genuine detection.
4) Classification Accuracy: Verifying the correctness of the assigned category for each detected object.

Of the various quality metrics, the assessment of completeness—that is, whether all relevant objects have been detected—is the most challenging aspect in the domain of P\&ID analysis. The dense, complex, and intricate nature of these diagrams makes it difficult to ascertain if any symbols have been missed. All the other aspects of quality assessment, while still important, can be addressed using analysis techniques in the localized regions of the bounding boxes of the detected objects. However, evaluating completeness requires a holistic understanding of the entire diagram on the one hand and ability to analyze local regions on the other for any symbols that should be present but are absent from the detection results. 

\section{Solution Approach}
\label{sec:solution}
 
\subsection{VLM Capabilities and Limitations}
Visual Language Models (VLMs) combine an image encoder and a text encoder that tokenize the input and produce embeddings.  A projection head maps both visual and textual embeddings into a shared high-dimensional space, enabling comparison and interaction. The fused embeddings are passed to a decoder to produce text conditioned on both visual and language inputs. Attention mechanisms are central to this process. Cross-attention links image patches to relevant text tokens, allowing the model to ground language in visual regions. Self-attention within each encoder captures relationships within a modality. This dual-attention setup enables VLMs dynamically align and integrate information across modalities.

\begin{figure}[!htbp]
  \centering
  \includegraphics[width=\columnwidth]{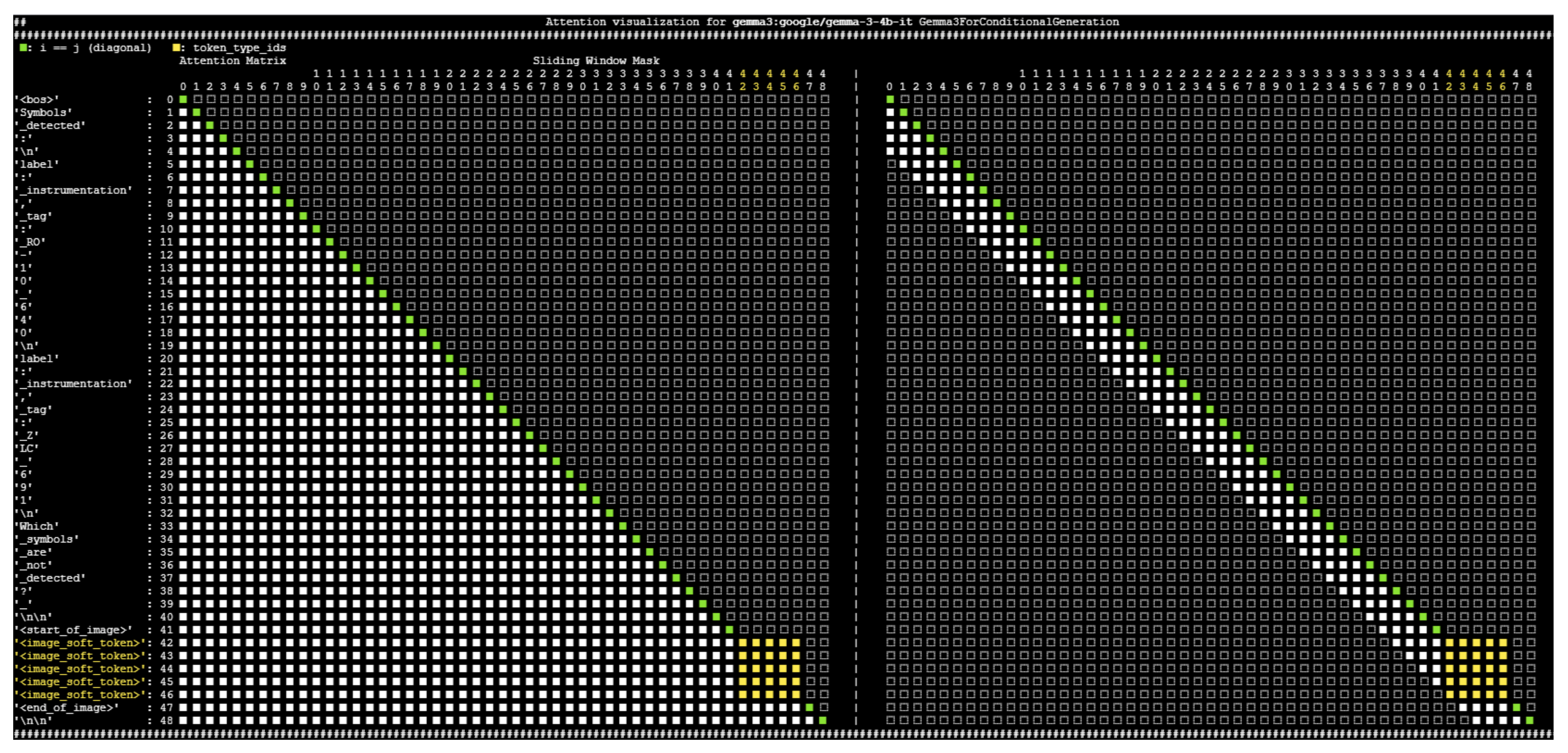}
  \caption{Attention Visualization.}
  \label{fig:pic4}
\end{figure}
Because of this architecture, VLMs are strong at high-level semantic reasoning, identifying objects, describing scenes, and interpreting contextual relationships. However, they remain limited in tasks that require precise visual or spatial reasoning and fine-grained geometric understanding. In the context of industrial diagrams, VLMs often struggle since the images are dense and cluttered  with many objects of varying scale and text. Moreover, the way the prompts are structured with the positioning of text and images can significantly impact the model's ability to focus on relevant regions of the image. Figure~\ref{fig:pic4} illustrates the attention patterns between tokens in the input text and image, highlighting the importance of carefully designing the prompt to elicit effective multimodal interactions. Some VLMs are trained for object detection. However, they still struggle to detect all the symbols in a P\&ID with high accuracy. This limitation is particularly pronounced in dense areas where multiple symbols are closely packed together.

\subsection{Proposed Framework for Object Detection Quality Assessment}
\label{ssec:agentic}

 \begin{figure*}[t!]
  \centering
  \includegraphics[width=\textwidth]{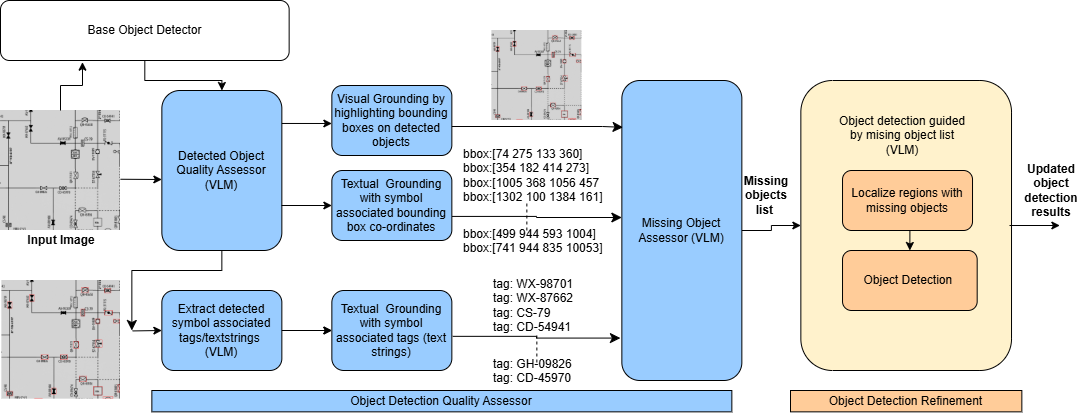}
  \caption{Proposed Framework for Object Detection Quality Assessment and Correction.}
  \label{fig:framework}
\end{figure*}
Figure~\ref{fig:framework} illustrates the proposed framework for object detection quality assessment and correction. The framework consists of two main components: 1) Object Detection Quality Assessor and 2) Object Detection Refinement. The base object detector can be any state-of-the-art object detector for P\&IDs.
The outputs of the object detector are first passed to the object detection quality assessor. The assessor evaluates the quality of the object detection results based on the four aspects mentioned in Section \ref{sec:problem}. The localized regions of the detected objects are first assessed for bounding box tightness, false detections and classification accuracy using a VLM in the detected object quality assessor. The good bounding boxes are filtered out at this stage and sent to subsequent stages.

In order to assess the outputs of the object detector for missing objects, the multi-modal capabilities of VLMs are leveraged along with the ability to perform high level semantic reasoning. Instead of using the VLM to directly detect objects/missing objects, it is used as a judge to assess the entire P\&ID for missing objects. Since the P\&ID is large, crops are taken and each crop is analyzed using the VLM. Even though the VLM is not expected to detect objects, it is still however required to analyze specific regions of the image – a capability that it is not inherently strong at without appropriate guidance or grounding. The proposed approach guides the VLM to focus on regions while keeping a global view of the image and context by different grounding mechanisms.   

To provide this necessary grounding, following methods were found to be effective:

1) Visual Grounding
To guide the VLM’s attention to specific image regions, visual cues can be provided. Two potential approaches are:
A) Use colored bounding boxes around detected objects and reference them in the prompt (e.g., drawing a red box around a valve and provide textual context, “The image has a valve in the red box”). However, this can become cluttered with many objects and depends on the VLM correctly interpreting the color–text association.
B) Remove detected objects and show only the remaining image, prompting the VLM to identify what might be missing. This can be challenging if objects cannot be cleanly removed without leaving artifacts and may lead to loss of contextual information.

2) Direct Textual Grounding: Instead of providing the VLM an image and asking, "Did anything get missed?",  the detected symbol associated tag can be provided as a list of text strings (e.g., "WX-98701", "CD-54941). This provides additional grounding for the VLM since this text is present in the image. To achieve this, the text strings corresponding to the symbols are first extracted as anchors. If the text strings are already extracted and associated with the symbols, depending on the certainty of this result, the symbol associated text string extraction could be skipped.These text strings are then provided to the VLM along with the image.
However, the challenge here is that some symbols might not have explicit tags. 

3) Explicit Geometric Grounding: Here the VLM  is provided the bounding box co-ordinates of the detected symbols in text format.  However, this approach is effective if the VLM natively understands bounding box co-ordinates. If the VLM does not have this capability, special tokens can be defined and the VLM can be fine-tuned to learn the embeddings for these new tokens.

4) Combination of the above approaches
Using a combination of the above approaches would result in giving as many cues to the VLM as possible to make it focus on the relevant regions of the image while keeping a global view of the image and context.

Based on the outcome of the object detection quality assessment, in the object detection refinement stage, each region is localized and a VLM based object detector is used to detect the missing object(s) in the region. The detected objects are then added to the object detection output to arrive at the final object detection output. 
\begin{table*}[ht]
\centering
\caption{Missing Object Assessor Performance}
\label{tab:results}
\begin{adjustbox}{max width=\textwidth}
\begin{tabular}{
l
S S S
S S S
S S S
}
\toprule
\multirow{2}{*}{Model} &
\multicolumn{3}{c}{Visual Grounding + Text Tags} &
\multicolumn{3}{c}{Visual Grounding + BBox Co-ordinates} &
\multicolumn{3}{c}{Visual Grounding + Text tags + BBox Co-ords} \\
\cmidrule(lr){2-4}\cmidrule(lr){5-7}\cmidrule(lr){8-10}
 & {Precision} & {Recall} & {F1} 
 & {Precision} & {Recall} & {F1}
 & {Precision} & {Recall} & {F1} \\
\midrule
Gemini 2.5         & 0.771 & 0.862 & 0.813 & 0.738 & 0.879 & 0.802 & 0.833 & 0.924 & 0.876 \\
Gemma3             & 0.755 & 0.804 & 0.778 & 0.706 & 0.832 & 0.763 & 0.761 & 0.879 & 0.815 \\
\bottomrule
\end{tabular}
\end{adjustbox}
\end{table*}
\section{Experiments}
\label{sec:experiments}

\subsection{Data Preparation and Metrics}
\label{ssec:dataset}
The 'Dataset PID' from the PID2GRAPH dataset \cite{sturmer_2025_14803338} has been used. It comprises 500 P\&IDs each of size 7168x4561 pixels. To generate a baseline dataset for the experiments, various strategies were employed to simulate common object detection errors. These included randomly omitting certain bounding boxes to mimic missed detections, altering bounding box dimensions, offsetting them and introducing false positives. This dataset allowed for a controlled evaluation of the VLM's performance in identifying and correcting these errors.

In order to evaluate the performance of the object detection quality assessor, following metrics are used.
\begin{itemize}
  \item VLM Judge Precision: $ \frac{TP}{TP+FP} $ — how many of the VLM's "missed" claims were actually correct
  \item VLM Judge Recall: $ \frac{TP}{TP+FN} $ — out of all missed objects, how many the VLM correctly identified
  \item VLM Judge F1 Score: $ \frac{2 \times (\text{Precision} \times \text{Recall})}{\text{Precision}+\text{Recall}} $ — provides a balanced measure of both precision and recall
\end{itemize}
\noindent
where, TP (true positives): VLM correctly identifies a missed ground-truth object; 
FP (false positives): VLM claims a missing object that either doesn't exist or was already detected; 
FN (false negatives): VLM fails to identify a missed object.
To compare the results of the object detection after correction mAP (mean Average Precision) is used. \cite{padilla} is used for calculating mAP.

\subsection{Implementation Details}
\label{ssec:implementation}
For VLM as a judge, closed model Gemini 2.5 Flash \cite{gemini} and open-weight model Gemma3 \cite{gemmateam2024gemmaopenmodelsbased} \cite{gemmateam2025gemma3technicalreport} have been used. For subsequent object detection for correction, PaliGemma \cite{PaliGemma,paligemma2} is used. The dataset described above comprises of around 6 million input tokens for Gemini 2.5 Flash model for one pass through the dataset. The current approach is a framework that leverages inference time capabilities of VLMs and does not involve any training or fine-tuning of the models in the experiments conducted. Depending on the complexity and nature of the diagrams, there might be some cases where the model might need to be fine-tuned. 

\subsection{Results and Analysis}
\label{ssec:results}
Table \ref{tab:results} shows the results of the VLM as a judge under different conditions to assess object detection quality, before any correction is done. Various grounding strategies were tried. Visual grounding using only highlighting the detected bounding boxes with a color like red  along with an elaborate text prompt inquiring about missed objects results in extremely poor recall of 0.32 (not reported in Table 1). This is expected since the VLMs lack precise spatial reasoning skills. When visual grounding is augmented with text tags corresponding to the detected symbols, the performance improves significantly. This is because the text tags provide additional semantic context that helps the VLM focus its attention on relevant areas of the image. There is some improvement in the recall, but reduction in precision when the visual grounding is augmented with the bounding box co-ordinates of the detected objects as compared to augmentation with text tags. However, this is only effective if the VLM natively understands bounding box co-ordinates. The best performance is observed when visual grounding is augmented with both the text tags and  bounding box co-ordinates. 
Moreover, as expected the larger model, Gemini, performs better than Gemma3. Gemma3, though being a smaller model has a reasonable recall implying missed object are identified reasonably well, however, the precision being poor indicates there are higher false positives. This can be overcome by a chain of thought prompting and the subsequent object detection stage, where it discards recommendations with no valid symbol.  Gemma3  as an open-weight model can be a highly capable candidate especially in scenarios where self-hosted solutions are required due to confidentiality reasons.
\begin{table}[ht]
\centering
\caption{System Performance}
\label{tab:objdet}
\resizebox{\columnwidth}{!}{%
\begin{tabular}{lcc}
\toprule
Model & mAP before correction & mAP after correction \\
\midrule
Gemini 2.5 & 0.728 & 0.925 \\
Gemma 3 & 0.728 & 0.882 \\
\bottomrule
\end{tabular}
} % end resizebox
\end{table}
Table \ref{tab:objdet} shows the results of the object detection after correction based on the outputs of the object detection quality assessor. The best performing quality assessor was used and results obtained for Gemini 2.5 Flash and Gemma 3. Since the object detector used for correction works on a localized region, it is quite accurate and is able to improve the mAP of the object detection results.

\section{Conclusion}
\label{sec:conclusion}

This paper has explored the application of VLM as automated judge for object detection in industrial diagrams, addressing the unique challenges posed by these complex and information-dense engineering entities. Within this complex quality assessment framework, VLMs are employed for multiple tasks, including symbol-associated tag extraction, bounding box localization accuracy assessment, identifying missing objects, and their subsequent detection. By leveraging VLMs' multimodal capabilities in different ways, it was demonstrated how quality assessment of object detection can be automated. The implications of this approach are significant as it not only reduces human workload in verifying object detection results, but also enables the development of agentic workflows that can adaptively respond to varying input conditions of the diagrams which enables robust scalability in industrial settings.

\vfill\pagebreak

% \section{REFERENCES}
% \label{sec:refs}
\bibliographystyle{IEEEbib}
\bibliography{refs}

\end{document}